\pdfoutput=1

\documentclass[11pt]{article}

\usepackage{ACL2023}

\usepackage{times}
\usepackage{latexsym}
\usepackage[T1]{fontenc}
\usepackage[utf8]{inputenc}
\usepackage{microtype}
\usepackage{enumitem}
\usepackage{inconsolata}
\usepackage{booktabs}
\usepackage{graphicx}
\usepackage{multirow}
\usepackage{dblfloatfix}
\usepackage{amsmath}
\usepackage{float}
\usepackage{adjustbox}
\usepackage{tabularx}
\usepackage{listings}
\usepackage[normalem]{ulem}
\usepackage{multicol}
\usepackage{lineno}
\usepackage{setspace}
\useunder{\uline}{\ul}{}

\newcommand{\textttt}[1]{%
  \begingroup
  \ttfamily
  \begingroup\lccode`~=`/\lowercase{\endgroup\def~}{/\discretionary{}{}{}}%
  \begingroup\lccode`~=`[\lowercase{\endgroup\def~}{[\discretionary{}{}{}}%
  \begingroup\lccode`~=`.\lowercase{\endgroup\def~}{.\discretionary{}{}{}}%
  \catcode`/=\active\catcode`[=\active\catcode`.=\active
  \scantokens{#1\noexpand}%
  \endgroup
}

\newcommand{\ie}{\textit{i}.\textit{e}., }
\newcommand{\eg}{\textit{e}.\textit{g}.\ }

\usepackage{xcolor}
\usepackage[many]{tcolorbox}

\newtcolorbox{boxI}{
    colback = sub, 
    colframe = main, 
    boxrule = 0pt, 
    toprule = 3pt
}
\definecolor{main}{HTML}{5989cf}
\definecolor{sub}{HTML}{cde4ff}

\title{Benchmarking LLMs on the Semantic Overlap Summarization Task}

\author{
 \textbf{John Salvador\textsuperscript{1}},
 \textbf{Naman Bansal},
 \textbf{Mousumi Akter\textsuperscript{2}},
 \textbf{Souvika Sarkar\textsuperscript{3}},
\\
 \textbf{Anupam Das\textsuperscript{4}},
 \textbf{Santu Karmaker\textsuperscript{1,2}}
\\
 \textsuperscript{1}{NLP@UCF, Department of Computer Science, University of Central Florida}\\
 \textsuperscript{2}{Research Center Trustworth Data Science and Security, Technical University of Dortmund}\\
 \textsuperscript{3}{School of Computing, Wichita State University}\\
 \textsuperscript{4}{Department of Computer Science, NC State University}
\\
 \small{
   \textbf{Correspondence:} \href{mailto:johnsalvador@ucf.edu}{johnsalvador@ucf.edu}, \href{mailto:santu@ucf.edu}{santu@ucf.edu}
 }
}

\begin{document}
\maketitle

%assessing their ability to summarize overlapping information from multiple alternative narratives

\begin{abstract}
Semantic Overlap Summarization (SOS) is a constrained multi-document summarization task, where the constraint is to capture the common/overlapping information between two alternative narratives. 
In this work, we perform a benchmarking study of popular Large Language Models (LLMs) exclusively on the SOS task. Additionally, we introduce the PrivacyPolicyPairs (3P) dataset to expand the space of SOS benchmarks in terms of quantity and variety.
This dataset provides 135 high-quality SOS data samples sourced from privacy policy documents. We then use a standard prompting taxonomy called TELeR to create and evaluate $905,216$ distinct LLM-generated summaries over two SOS datasets from different domains, and we further conduct human evaluation on a subset of 540 samples.
We conclude the paper by analyzing models' performances and the reliability of automatic evaluation\footnote{The code and datasets used to conduct this study are available at \url{https://anonymous.4open.science/r/llm_eval-E16D}}.
\end{abstract}

%We also introduce a novel evaluation dataset for the SOS task, which we call the \textit{PrivacyPolicyPairs} (3P) dataset, facilitating robust experimentation and benchmarking. , an alternate domain of text from the original SOS dataset

%When evaluating LLMs on these datasets prompt engineering is vital due to the fact that performance is highly dependent on prompt quality.So then to maximize our potential benchmark scores we employ the TELeR taxonomy \cite{teler}, which outlines categories of prompting methods for instruction-tuned LLMs.

\section{Introduction}

%Large Language Models (LLMs) represent a groundbreaking advancement in the research landscape of Natural Language Processing (NLP) and Artificial Intelligence (AI). 
%Trained on large bodies of text data, LLMs excel in generating coherent and human-like text. 
%These models have been evaluated in a wide range of NLP tasks~\cite{bubeck2023sparks,dai2022can,du2022glam,smith2022using} across several areas, including software development, law, and medicine~\cite{llm_for_unit_test, llm_for_law, llm_for_medicine}. 
%However, there are still areas and tasks where LLMs are yet to be rigorously evaluated. 
In the field of Natural Language Processing (NLP), Large Language Models (LLMs) have proven themselves to be the most capable text generation models in a variety of tasks and fields \citep{bubeck2023sparks,dai2022can,du2022glam,smith2022using,llm_for_unit_test, llm_for_law, llm_for_medicine}.
One task where LLMs are understudied is Semantic Overlap Summarization (SOS)~\cite{semantic_overlap,sofsat}, where the goal is to summarize the common/overlapping information between two alternative narratives conveying similar information.
Applications for this task include isolating facts from opinions in news articles, aggregating consistent claims across legal or medical documents, and extracting common issues from user reviews. 
Such capabilities are especially well suited for LLMs, which can process long-context inputs and generate fact-based responses grounded in multiple sources. 
In this setting, SOS serves as a proxy for trust-aware summarization, where overlapping content can be used to strengthen citation quality and reduce hallucination in generation.
This is particularly important for applications where factual reliability and trust are paramount such as medical, legal, or journalistic contexts. since identifying and summarizing overlapping content across independent sources can serve as a proxy for information corroboration. SOS thus enables LLMs to produce outputs that are not only concise but also grounded in their multiple inputs, enhancing transparency and trustworthiness in generation.
%In this paper, we conduct a comprehensive benchmarking study on the SOS task using 16 popular LLMs. 
In this paper, we conduct a comprehensive benchmarking study on how LLMs perform on the SOS task using 16 popular models. 

As LLMs' performance can widely vary with prompt variations \cite{prompts_matter, prompt_sens_1}, we use a standard prompting taxonomy, TELeR~\cite{teler}, to devise a comprehensive set of prompts with different degrees of detail before invoking LLMs to perform the SOS task. Our evaluation includes two different alternative narrative-pairs datasets. 
The first dataset is the \textit{AllSides} dataset released by~\citet{semantic_overlap}, and the second dataset is our original contribution, which was built with extensive human annotation effort, which we name as the \textit{PrivacyPolicyPairs} (3P) dataset. 

%\input{fig/sos_for_privacy_policy}

%Figure~\ref{fig:sos_for_privacy_policy} illustrates an example of the SOS task when comparing two alternative privacy policies in the 3P dataset, where the green text denotes the output (common information) from two input privacy policies, one from Google and one from Apple. 
%In this case, we have two competing platforms that provide similar types of services (\eg Cloud Storage or Streaming Services) and each company lays out its practices when handling your private information. 
%These documents contain important information regarding your privacy but can be long and cumbersome to read. The SOS task can help users by briefly identifying the common practices followed by each company.

We report ROUGE, BERTscore, and SEM-$F_1$ on the Allsides and 3P datasets for each combination of LLMs and prompt style, totaling 905,216 distinct samples. We further collected human annotations on a subset of 540 samples to truly gauge the capabilities of LLMs in capturing overlapping information from multiple narratives.
Finally, we analyze LLMs' performances and the reliability of automatic evaluation via correlation analysis against human annotations.

\section{The Benchmark Datasets}\label{datasets}
%We experimented on two datasets designed specifically for the Semantic Overlap Summarization task. 

\subsection{The AllSides Data}\label{allsides}
%The AllSides dataset is collected from AllSides.com, a third-party online news forum known for presenting news and information from various political perspectives. 
The AllSides dataset is the first to be introduced for the SOS task. 
To build this dataset, \citet{semantic_overlap} crawled news articles from AllSides.com to create 2,788 sample training set and 137 sample test set.
Each sample contains 2 source documents of left and right-leaning sources and is accompanied by a reference summary.
The test set includes an additional 3 human-annotated summaries for more robust evaluation.

\subsection{The \textit{PrivacyPolicyPairs} (3P) Data} \label{privacypolicy}
For a more diverse evaluation, we introduce the \textit{PrivacyPolicyPairs} (3P) dataset, focusing on the SOS task for a different domain and containing $135$ human-annotated samples. 
Each sample comprises $2$ source documents (two different privacy policy narratives), the category of passage, 3 reference summaries, company names, and word counts (example figure in the appendix). 
%An example is shown in Table~\ref{tbl:3psample}. 
Our (3P) dataset is built on the OPP-115 Corpus introduced by \citet{privacy_policy_original}, which comprises $115$ privacy policies (267K words) spanning $15$ sectors (Arts, Shopping, News, etc.).
The policy data of the OPP-115 corpus are also tagged with the following categories:

\begin{multicols}{2}\small
\begin{itemize}[leftmargin=*,itemsep=0ex,partopsep=0ex,parsep=0ex]
    \item First Party Collection/Use 
    \item Third Party Sharing/Collection
    \item User Choice/Control
    \item User Access, Edit, \& Deletion
    \item Data Retention
    \nolinenumbers
    \item Data Security
    \item Policy Change
    \item Do Not Track
    \item International \& Specific Audiences
    \item Other
\end{itemize}
\end{multicols}
% \linenumbers

These categories are associated with text spans in each document that denote where the labels were relevant. Our motivation behind introducing a new dataset for SOS evaluation is to
1) extend the amount of available testing data from just 137 samples from the AllSides evaluation set to 272 total evaluation samples with a combined total of 953 human annotations and 
%2) The 3P dataset represents a new type of documents in the form of semi-structured privacy policies as opposed to the news articles that make up the AllSides data; 
%2) The 3P dataset is made up of privacy policy documents, a different domain from the AllSides data.
2) provide data from a domain different from the AllSides data.

\smallskip\noindent\textbf{Constructing the 3P Dataset:}\label{constructing_3p}
%The 3P dataset includes pairs of passages taken from the OPP-115 corpus and tasks the annotator with finding the semantically overlapping information between them. 
%The 3P dataset includes pairs of passages taken from the OPP-115 corpus. 
%Each sample comprises $2$ source documents (two alternative privacy policy narratives), the category they fall under, and $3$ reference overlap summaries. The company names and word counts are also included.
%A data sample is shown in Table~\ref{tbl:3psample}. 
%When curating this dataset, we wanted to ensure each passage pair had some degree of overlap. 
%To facilitate this goal, we reversed the process followed by the original authors and grouped the documents back into their respective sectors. 
To build the 3P dataset, we set out to create pairs of passages from the original OPP-115 corpus.
%To ensure that each document pair has some degree of overlap, we first grouped each document into the 15 sectors that were originally assigned by \citet{privacy_policy_original} (Arts, Shopping, Business, News, etc.). 
To ensure a degree of overlap, we first grouped each document into the 15 sectors that were originally assigned by \citet{privacy_policy_original} (Arts, Shopping, Business, News, etc.). 
Then, within each sector, we paired different passages according to their category labels (First Party Collection, Data Retention, etc.). 
This process resulted in $6110$ passage pairs across all sectors.
% \subsubsection{AllSides vs. 3P}
% The AllSides and 3P datasets are both used to evaluate a model's SOS capabilities.
% the 3P dataset features a 

%\input{fig/ds_stats}

Out of the 15 sectors, we focused on \textit{eCommerce}, \textit{Technology}, and \textit{Food and Drink}. %due to their popularity as well as diversity among each other. From these sectors, we collected $346$ passage pairs to annotate. 
We then recruited three volunteer annotators from the department and instructed them to write a summary of common information present in each document pair. 
The exact instructions can be found in Appendix \ref{human_annotation}.
After the initial round of annotation, the annotators came together, discussed the differences in each of their summaries, and revised their original summaries accordingly.
After revising and removing samples with no overlap, we yielded 3 annotations per passage pair for a total of 405 annotations for 135 high-quality samples.
%After this discussion the annotators then revised their summaries.
%The process of annotating, resolving conflicts, and removing samples with no overlap resulted in 3 annotations per passage pair for a total of 405 annotations for 135 high-quality samples. 
%The final dataset statistics are listed in Table~\ref{tbl:ds_stats}.

%\input{fig/max_scores_per_model_family}
\section{Methodology}

\subsection{Evaluated Large Language Models}
%The development of reinforcement learning with human feedback (RLHF) has allowed LLMs to set a new standard for generative language models, allowing users to interact with them as databases using natural language queries \cite{rlhf}. 
%With these capabilities in mind, 
We choose to test our datasets using 7 families of instruction-tuned LLMs, totaling 16 models which are listed in Table \ref{tbl:llm_families}. 
OpenAI and Google provide their own unique APIs but for open source LLMs, we used the transformers library \cite{huggingface} to access model weights and run inference on a server with  4 Nvidia A4500 20GB GPUs.  
For additional speedup, we utilized the vLLM library \cite{vllm}.

% \usepackage{booktabs}
% \usepackage{multirow}
% \usepackage{graphicx}
\begin{table}[!htb]
\centering
\resizebox{0.9\linewidth}{!}{%
\begin{tabular}{ll}
\hline
\multicolumn{1}{c}{LLM Family} & \multicolumn{1}{c}{Model} \\
\hline\hline
Google Gemini & gemini-1.5-pro-001 (May 2024) \\       \cite{gemini} & \\\hline
OpenAI     & gpt-3.5-turbo-0125 (May 2024)\\
\cite{gpt4} & \\\hline
& mosaicml/mpt-7b-chat (7B) \\
MosaicML MPT                   & mosaicml/mpt-30b-chat (30B) \\
\cite{mpt}        & mosaicml/mpt-7b-instruct (7B) \\
& mosaicml/mpt-30b-instruct (30B)            \\\hline
& lmsys/vicuna-7b-v1.5 (7B)\\
{LMSYS Vicuna}  & lmsys/vicuna-13b-v1.5 (13B) \\
\cite{vicuna} & lmsys/vicuna-7b-v1.5-16k (7B)\\
& lmsys/vicuna-13b-v1.5-16k (13B) \\\hline                    
{MistralAI}    & mistralai/Mistral-7B-Instruct-v0.1 (7B)\\
\cite{mistral}  & mistralai/Mistral-7B-Instruct-v0.2 (7B) \\\hline
{MetaAI Llama2} & meta-llama/Llama-2-7b-chat-hf (7B)\\
\cite{llama2} & meta-llama/Llama-2-13b-chat-hf (13B)\\\hline
{Microsoft Phi-3} & microsoft/Phi-3-mini-4k-instruct (3.8B)\\
\cite{phi3} & microsoft/Phi-3-mini-128k-instruct 3.8B)\\\hline
\end{tabular}%
}\vspace{-2mm}
\caption{
   The list of models evaluated in this paper with parameter counts. 
   We use $7$ families of models, $2$ of which are closed source, and $5$ open source. 
   %Known parameter count models are included in parentheses.
   %OpenAI and Google have not reported the parameter counts of their models.
}\vspace{-4mm}
\label{tbl:llm_families}
\end{table}

\subsection{Prompt Design}
\label{sec:designed_prompts}
%For example, works from~\citet{zero_shot_topic_inf, zero_shot_cls} explore their zero-shot use cases in topic inference and text classification.
%We chose not to prompt using levels 5 and 6 because their use of retrieval augmented prompting does not necessarily apply to the SOS task due to all relevant context being present, \ie the two source narratives are already provided as part of the prompt. 
%Furthermore, requirement number 5 for level 6 also specifies asking the LLM to explain its own output, which would negatively affect the generated summaries during evaluation. We also experiment with in-context learning prompts~\cite{gpt3}.
%For each template, we use the following outline for our prompt design.

%Simply following a general prompting taxonomy does not guarantee the highest evaluation scores.
%So then to 

% To ensure comprehensive prompt engineering, we created $5$ variations for each TELeR level~\cite{teler} from level 0 to level 4, i.e., we created $5\times 5=25$ different unique prompts. 
% To avoid confusion, we label these variations A to E. 
% Each variation set follows a similar layout to what was previously mentioned but with variations in word choice, directive styles, etc.
%We prompted LLMs in a zero-shot setting using the TELeR taxonomy as zero-shot approaches to NLP tasks have gained popularity with the growing capabilities of LLMs~\cite{zero_shot_topic_inf, zero_shot_cls}. 
We prompted LLMs in a zero-shot setting as these methods have gained popularity with the growing capabilities of LLMs~\cite{zero_shot_topic_inf, zero_shot_cls}. 
Specifically, we utilize the guidelines laid out by the TELeR taxonomy due to its use and reference in previous studies \cite{teler_sup1, teler_sup2, teler_sup3, teler_sup4, teler_sup5, teler_sup6}.
For this study, we used TELeR levels 0 through 4 (5 out of the 7). 
To ensure comprehensive prompt engineering, we created templates for TELeR levels 0 through 4 and In-Context Learning styled prompts \cite{gpt3} (details in appendix~\ref{appendix:prompt_design}).
 %In each template group we create prompting text that can be used or only the AllSides data, only the 3P data or for both.
For each template, we then created variations of prompts that follow their respective formats.
For example, the group of TELeR L1 prompts is comprised of 8 prompts: 5 general, 3 AllSides-specific, and 3 3P-specific.
Then, to construct our final set of prompts, we took all possible combinations of system roles and prompts, creating {$\mathbf{56,576}$} prompts for each of our 16 models and, thus, creating $\mathbf{905,216}$ distinct evaluation samples in total.
%Using this prompting strategy, we've created 56,576 unique prompts for each of our $16$ evaluated LLMs, totaling 905,216 evaluation samples.
%More details are provided in appendix~\ref{appendix:prompt_design}.
%A breakdown of the variation counts for each group is shown in Table~\ref{tbl:prompt_variations}.
%See appendix~\ref{appendix:prompt_design} for the exact prompts that were used.
 
%#To avoid confusion, we label these variations A to E. 
%Each variation set follows a similar layout to what was previously mentioned but with variations in word choice, directive styles, etc.

%After designing our prompts, we evaluated them using $16$ LLMs for both datasets. 
%To ensure fairness, we report results only for the set of prompts with the highest average score for each metric across all the evaluated LLMs. 
%Table~\ref{tbl:prompt_group_trials} shows that prompt group A scored the highest averages across all metrics, so we used that set for final evaluations.

%\input{fig/prompt_variations}
%\input{fig/prompt_group_trials}

%\santu{You need to provide some justification as to why levels 5 and 6 were not tried.}

\newcolumntype{C}{>{\centering\arraybackslash}p{4.5em}}
\newcommand\boldblue[1]{\textcolor{blue}{\textbf{#1}}}

\begin{table*}[!htb] \LARGE
\begin{singlespace}\vspace{-2mm}
\centering
\resizebox{\textwidth}{!}{%
%\begin{tabular}{lcccccccc}
\begin{tabular}{lCCCCCCCCCCCCC}
 & \multicolumn{1}{l}{} & \multicolumn{1}{l}{} & \multicolumn{1}{l}{} & \multicolumn{1}{l}{} & \multicolumn{1}{l}{} & \multicolumn{1}{l}{} & \multicolumn{1}{l}{} & \multicolumn{1}{l}{} \\ \hline
\multicolumn{14}{c}{\textbf{AllSides Dataset}} \\ \toprule\toprule
%\multicolumn{1}{l|}{Model} & Sem-F1 (USE) & Sem-F1 (Distil) & Sem-F1 (RoB) & R-L Sum & R-L & R-1 & R-2 & \multicolumn{1}{c}{BERTscore} \\ \hline
% remove 1, 3
\multicolumn{1}{l|}{Model} & R-L Sum  & R-L  & R-1  & R-2  & BLEU  & METEOR  & chrF  & TER $\downarrow$ & Sem-F1  & \multicolumn{1}{c}{\begin{tabular}[c]{@{}c@{}}BERT\\ score\end{tabular}}  & BLEURT  & \multicolumn{1}{c}{\begin{tabular}[c]{@{}c@{}}Mover\\ score\end{tabular}}  & SMS  \\ \hline
\multicolumn{1}{l|}{gemini-pro} & 0.418 (l1) & 0.418 (l1) & \textbf{0.499} (l1) & \textbf{0.331} (l1) & 0.003 (l0)  & \textbf{0.538} (l1) & \boldblue{54.634} (l1) & 138.21 (l1) & \textbf{0.643} (l1) & \boldblue{0.503} (l1) & \boldblue{-0.144} (l1) & \boldblue{0.617} (l1) & \boldblue{0.617} (l1)\\ \hline
\multicolumn{1}{l|}{gpt-3.5-turbo} & 0.421 (l1) & 0.421 (l1) & 0.494 (l1) & 0.300 (l1) & 0.003 (icl) & 0.528 (l1)  & 53.151 (l1) & 148.21 (l1) & 0.641 (l4) & 0.490 (l1) & \textbf{-0.174} (l1) & \textbf{0.616} (l1) & 0.612 (l1)\\ \hline
\multicolumn{1}{l|}{vicuna-13b-v1.5} & 0.330 (l3) & 0.317 (l3) & 0.426 (l2) & 0.231 (l2) & 0.004 (l1) & 0.487 (l2) & 49.272 (l2) & 142.27 (l2) & 0.528 (l2) & 0.393 (l2) & -0.412 (l3) & 0.586 (l3) & 0.590 (l1)\\
\multicolumn{1}{l|}{vicuna-13b-v1.5-16k} & 0.326 (l2) & 0.296 (l4) & 0.410 (l2) & 0.236 (l1) & 0.003 (l1) & 0.462 (l3) & 47.970 (l1) & \textbf{130.22} (l1) & 0.535 (l3) & 0.362 (l2) & -0.440 (l3) & 0.581 (l4) & 0.590 (l1)\\
\multicolumn{1}{l|}{vicuna-7b-v1.5} & 0.355 (l2) & 0.333 (l2) & 0.446 (l2) & 0.255 (l2) & 0.004 (l1) & 0.497 (l2) & 50.817 (l2) & 321.37 (l3) & 0.549 (l4) & 0.405 (l2) & -0.439 (l3) & 0.590 (l2) & 0.595 (l2)\\
\multicolumn{1}{l|}{vicuna-7b-v1.5-16k} & 0.323 (l2) & 0.309 (l2) & 0.419 (l2) & 0.231 (l2) & 0.004 (l1) & 0.484 (l2) & 48.843 (l2) & 308.47 (l3) & 0.550 (l3) & 0.387 (l2) & -0.407 (l3) & 0.582 (l2) & 0.586 (l2)\\ \hline
\multicolumn{1}{l|}{Llama-2-13b-chat-hf} & 0.372 (l1) & 0.357 (l1) & 0.442 (l1) & 0.257 (l1) & 0.002 (l1) & 0.495 (l4) & 49.459 (l1) & 236.98 (l1) & 0.563 (l2) & 0.369 (l1) & -0.468 (l2) & 0.592 (l1) & 0.584 (l1)\\ 
\multicolumn{1}{l|}{Llama-2-7b-chat-hf} & 0.336 (l3) & 0.332 (l1) & 0.434 (l3) & 0.239 (l3) & 0.002 (l1) & 0.498 (l3) & 49.593 (l3) & 251.37 (l4) & 0.603 (l2) & 0.402 (l3) & -0.309 (l1) & 0.589 (l1) & 0.588 (l3) \\ \hline
\multicolumn{1}{l|}{Phi-3-mini-128k-instruct} & { \boldblue{0.442} (l3)} & { \boldblue{0.433} (l3)} & { \boldblue{0.507} (l3)} & { \boldblue{0.342} (l3)} & 0.003 (l2) & \boldblue{0.541} (l1) & \textbf{54.296} (l1) & 156.74 (l2) & \boldblue{0.646} (l1) & 0.480 (l1) & -0.179 (l1) & \textbf{0.616} (l1) & \textbf{0.623} (l1)\\
\multicolumn{1}{l|}{Phi-3-mini-4k-instruct} & 0.375 (l1) & 0.375 (l1) & 0.453 (l1) & 0.255 (l1) & 0.002 (l1) & 0.493 (l1) & 49.756 (l1) & 198.04 (l1) & 0.607 (l3) & 0.445 (l1) & -0.188 (l1) & 0.600 (l1) & 0.588 (l1)\\ \hline
\multicolumn{1}{l|}{Mistral-7B-Instruct-v0.1} & \textbf{0.428} (l1) & \textbf{0.428} (l1) & 0.498 (l1) & 0.318 (l1) & 0.002 (l1) & 0.539 (l1) & 53.128 (l1) & 190.72 (l1) & 0.636 (l3) & \textbf{0.494} (l1) & -0.194 (l1) & 0.614 (l1) & 0.613 (l1)\\
\multicolumn{1}{l|}{Mistral-7B-Instruct-v0.2} & 0.374 (l1) & 0.374 (l1) & 0.464 (l1) & 0.268 (l4) & 0.002 (l0) & 0.511 (l4) & 51.546 (l4) & 253.57 (l1) & 0.637 (l1) & 0.462 (l1) & -0.229 (l1) & 0.601 (l1) & 0.596 (l1)\\ \hline
\multicolumn{1}{l|}{mpt-30b-chat} & 0.340 (l1) & 0.338 (l1) & 0.419 (l1) & 0.252 (l1) & 0.001 (l2) & 0.476 (l2) & 47.994 (l2) & 520.50 (l2) & 0.596 (l1) & 0.374 (l2) & -0.319 (l2) & 0.588 (l1) & 0.591 (l1) \\
\multicolumn{1}{l|}{mpt-30b-instruct} & 0.345 (l1) & 0.345 (l1) & 0.427 (l1) & 0.237 (l2) & \boldblue{0.010} (l3) & 0.445 (l2) & 46.618 (l2) & \boldblue{112.52} (l2) & 0.602 (l2) & 0.435 (l1) & -0.309 (l1) & 0.593 (l1) & 0.588 (l2)\\
\multicolumn{1}{l|}{mpt-7b-chat} & 0.267 (l4) & 0.263 (l3) & 0.356 (l3) & 0.206 (l4) & 0.003 (icl) & 0.434 (l4) & 43.745 (l4) & 327.89 (l2) & 0.578 (l4) & 0.304 (l3) & -0.378 (l3) & 0.593 (l2) & 0.585 (l4)\\
\multicolumn{1}{l|}{mpt-7b-instruct} & 0.278 (l1) & 0.277 (l1) & 0.370 (l1) & 0.195 (l1) & \textbf{0.006} (l4) & 0.422 (l1) & 44.214 (l1) & 134.32 (l4) & 0.585 (l2) & 0.316 (l3) & -0.378 (l3) & 0.571 (l1) & 0.586 (l2)\\ \midrule\midrule

\multicolumn{14}{c}{\textbf{PrivacyPolicyPairs (3P) Dataset}} \\ \hline\hline
\multicolumn{1}{l|}{gemini-pro} & \textbf{0.244} (l4) & \textbf{0.243} (l4) & \textbf{0.314} (l4) & \boldblue{0.118} (l1) & 0.003 (icl) & \textbf{0.347} (l4) & \boldblue{41.843} (l4) & \textbf{150.77} (l1) & \textbf{0.528} (l4) & \textbf{0.308} (l1) & \textbf{-0.198} (l2) & \textbf{0.561} (l4) & \textbf{0.545} (l4)\\ \hline
\multicolumn{1}{l|}{gpt-3.5-turbo} & \boldblue{0.262} (l1) & \boldblue{0.262} (l1) & \boldblue{0.324} (l1) & \textbf{0.117} (l1) & 0.003 (l1) & \boldblue{0.355} (l1) & \textbf{41.186} (l2) & 171.67 (l1) & \boldblue{0.535} (l4) & \boldblue{0.329} (l1) & \boldblue{-0.156} (l1) & \boldblue{0.567} (l1) & \boldblue{0.546} (l1)\\ \hline
\multicolumn{1}{l|}{vicuna-13b-v1.5} & 0.196 (l2) & 0.180 (l2) & 0.250 (l2) & 0.088 (l2) & 0.002 (l2) & 0.339 (l2) & 37.375 (l2) & 322.60 (l2) & 0.445 (l3) & 0.205 (l2) & -0.463 (l4) & 0.552 (l3) & 0.533 (l2)\\
\multicolumn{1}{l|}{vicuna-13b-v1.5-16k} & 0.184 (l2) & 0.171 (l2) & 0.239 (l2) & 0.077 (l2) & 0.003 (l1) & 0.318 (l2) & 36.181 (l2) & 164.16 (l1) & 0.471 (l0) & 0.189 (l2) & -0.423 (l4) & 0.546 (l2) & 0.529 (l2)\\
\multicolumn{1}{l|}{vicuna-7b-v1.5} & 0.175 (l2) & 0.165 (l2) & 0.227 (l2) & 0.071 (l2) &  0.005 (l1) & 0.308 (l2) & 35.699 (l2) & 460.12 (l1) & 0.441 (l4) & 0.177 (l2) & -0.501 (l1) & 0.543 (l1) & 0.527 (l1)\\
\multicolumn{1}{l|}{vicuna-7b-v1.5-16k} & 0.188 (l1) & 0.186 (l1) & 0.247 (l1) & 0.069 (l2) & 0.003 (l1) & 0.303 (l3) & 36.652 (l1) & 375.69 (l1) & 0.497 (l3) & 0.204 (l1) & -0.404 (l4) & 0.553 (l3) & 0.533 (l3)\\ \hline
\multicolumn{1}{l|}{Llama-2-13b-chat-hf} & 0.207 (l1) & 0.196 (l1) & 0.266 (l1) & 0.083 (l1) & 0.001 (l1) & 0.305 (l1) & 38.272 (l1) & 340.60 (l1) & 0.466 (l3) & 0.184 (l1) & -0.500 (l4) & 0.545 (l1) & 0.531 (l1)\\
\multicolumn{1}{l|}{Llama-2-7b-chat-hf} & 0.199 (l1) & 0.197 (l1) & 0.258 (l1) & 0.079 (l1) & 0.001 (l1) & 0.300 (l4) & 37.899 (l1) & 361.54 (l1) & 0.495 (l1) & 0.214 (l1) & -0.383 (l1) & 0.547 (l1) & 0.529 (l1)\\ \hline
\multicolumn{1}{l|}{Phi-3-mini-128k-instruct} & 0.218 (l3) & 0.217 (l3) & 0.282 (l3) & 0.083 (l1) & 0.003 (l4) & 0.308 (l1) & 37.816 (l1) & 187.90 (l4) & 0.497 (l1) & 0.276 (l1) & -0.205 (l1) & 0.554 (l1) & 0.533 (l1)\\
\multicolumn{1}{l|}{Phi-3-mini-4k-instruct} & 0.215 (l1) & 0.215 (l1) & 0.278 (l1) & 0.083 (l1) & 0.002 (l1) & 0.321 (l1) & 38.572 (l1) & 259.86 (l1) & 0.503 (l1) & 0.251 (l1) & -0.345 (l1) & 0.551 (l1) & 0.529 (l1)\\ \hline
\multicolumn{1}{l|}{Mistral-7B-Instruct-v0.1} & 0.214 (l1) & 0.213 (l1) & 0.275 (l1) & 0.083 (l1) & 0.002 (l1) & 0.330 (l1) & 37.823 (l4) & 238.45 (l1) & 0.517 (l1) & 0.249 (l1) & -0.362 (l2) & 0.549 (l1) & 0.535 (l1)\\
\multicolumn{1}{l|}{Mistral-7B-Instruct-v0.2} & { 0.234 (l1)} & { 0.233 (l1)} & { 0.298 (l1)} & { 0.106 (l1)} & 0.002 (l1) & 0.340 (l4) & 39.959 (l1) & 247.36 (l1) & 0.523 (l1) & 0.279 (l1) & -0.291 (l1) & 0.558 (l1) & 0.540 (l1)\\ \hline
\multicolumn{1}{l|}{mpt-30b-chat} & 0.192 (l1) & 0.190 (l1) & 0.247 (l1) & 0.075 (l1) & 0.002 (l1) & 0.312 (l2) & 35.142 (l2) & 385.01 (l2) & 0.507 (l2) & 0.200 (l2) & -0.347 (l2) & 0.655 (icl) & 0.534 (l2)\\
\multicolumn{1}{l|}{mpt-30b-instruct} & 0.213 (l1) & 0.210 (l1) & 0.267 (l1) & 0.084 (l1) & \boldblue{0.014} (l1) & 0.297 (l2) & 35.520 (l1) & \boldblue{131.85} (l1) & 0.487 (l2) & 0.268 (l1) & -0.361 (l1) & 0.667 (icl) & 0.538 (l1)\\
\multicolumn{1}{l|}{mpt-7b-chat} & 0.177 (l2) & 0.175 (l2) & 0.233 (l2) & 0.066 (l1) & 0.003 (l0) & 0.270 (l1) & 33.066 (l2) & 352.14 (l2) & 0.479 (l2) & 0.159 (l2) & -0.464 (l3) & 0.651 (icl) & 0.530 (l1)\\
\multicolumn{1}{l|}{mpt-7b-instruct} & 0.166 (l1) & 0.162 (l1) & 0.215 (l1) & 0.075 (l1) & \textbf{0.006} (l4) & 0.270 (l2) & 33.105 (l1) & 152.96 (l4) & 0.469 (l1) & 0.127 (l1) & -0.561 (l1) & 0.654 (icl) & 0.529 (l1)\\ \hline
\end{tabular}%
}

\end{singlespace}
\caption{
    The best average scores for each metric over each dataset. Higher is better for all but TER which is indicated by $\downarrow$.
    Bold blue indicates the best score for a given metric, while the second best is indicated by bold black.
    Each score is accompanied by the TELeR level that was used to produce the score.
}
\label{tbl:best_scores}
\end{table*}
\subsection{Evaluation}
%We evaluate on $3$ types of metrics and $8$ in total: ROUGE \cite{rouge}, BERTscore \cite{bertscore}, and Sem-F1 \cite{semf1}.

% We conduct automatic evaluation using ROUGE~\citep{rouge}, BERTscore~\citep{bertscore}, and SEM-F1~\citep{semf1}. 
% SEM-F1 is an embedding-based metric like BERTscore but utilizes sentence-level embeddings to calculate its scores compared to BERTscore's use of token embeddings\footnote{We calculate BERTscore with hashcode roberta-large\_L17\_no-idf\_version=0.3.12(hug\_trans=4.40.2)-rescaled.
% For SEM-F1, we compute scores using 3 different models: USE\cite{use}, RoBERTa \cite{roberta}, and DistilRoBERTa \cite{distilbert}}.

\smallskip\noindent\textbf{Automatic Evaluation:} We conduct automatic evaluation using 11 different metrics.
For lexical overlap metrics we use \textbf{ROUGE}~\citep{rouge}, \textbf{BLEU}~\citep{bleu}, \textbf{METEOR}~\citep{meteor}, \textbf{chrF}~\citep{chrf}, \textbf{Translation Edit Rate}~\citep{ter}, 
and \textbf{CIDEr}~\citep{cider}. 
For embedding-based metrics we use \textbf{BERTscore}~\citep{bertscore}, \textbf{SEM-F1}~\citep{semf1}, \textbf{BLEURT}~\citep{bleurt}, \textbf{MoverScore}~\citep{moverscore}, and \textbf{Sentence Mover's Similarity}~\citep{sms}. See Appendix~\ref{appendix:eval} for details of each metric.

% \usepackage{graphicx}
%\begin{tabular}[c]{@{}c@{}}\textbf{\#} \textbf{Top} \\ \textbf{Scores}\end{tabular}
\begin{table}[!htb]\small
\centering
\begin{tabular}{r|c||r|c}
\hline
%\multicolumn{1}{r|}{\textbf{Model}}    & \textbf{Score (0-5)} & \multicolumn{1}{r|}{\textbf{Template}} & \textbf{Score (0-5)} \\ \hline\hline
\multicolumn{1}{r|}{\textbf{Model}}     & \begin{tabular}[c]{@{}c@{}}\textbf{Score} \\ \textbf{(0-5)}\end{tabular}
& \multicolumn{1}{r|}{\textbf{Template}} & \begin{tabular}[c]{@{}c@{}}\textbf{Score} \\ \textbf{(0-5)}\end{tabular} \\ \hline\hline
gemini-pro                              & 3.37              & ICL   & 3.08 \\
gpt-3.5-turbo                           & \textbf{3.53}     & L1    & 3.38 \\
mpt-30b-chat                            & 3.39              & L2    & \textbf{3.42} \\
Mistral-7B-Instruct-v0.2                & 3.38              & L3    & 3.32 \\
Phi-3-mini-128k-instruct                & 3.37              & L4    & 3.32 \\
vicuna-13b-v1.5-16k                     & 3.32              &       & \\ \hline
\end{tabular}%

\caption{
    Average negotiated preference score for each model and prompt template.
    "ICL" represents the In-Context Learning style prompts, while "Lx" refers to the level of the TELeR prompt.
}
\label{tbl:human_prefs}
\end{table}

\smallskip\noindent\textbf{Human Evaluation:}\label{sec:human_evaluation} We recruited 3 human volunteer for annotation purposes.
To avoid the burden of having annotators analyze 9 million samples, we reduce the number of evaluation samples by 1) evaluating a subset of data that corresponds to 15 narrative pairs (7 from AllSides and 8 from 3P) out of the 272 test set samples from AllSides and 3P, 2) evaluating only the largest/newest models from each family and 3) evaluating only the summaries that correspond to the best-performing prompts within each TELeR level. 
%This strategy reduced the number of summaries per sample from 3,328 to 36 summaries per sample, giving us a total of 540 samples per annotator. 
This strategy reduced the number of summary evaluations from 9M to 540 samples per annotator.
%The annotators were tasked to score the summaries on a scale of 0-5 based on how well they captured the overlapping information of the 2 given source documents. 
The annotators scored model summaries on a scale of 0-5 based on how well they captured the overlapping information between the two documents given. After individually scoring the summaries, the annotators sat together to resolve disagreements and assign a final score to each sample, giving us 2,160 scores across all samples.

\section{Results}

\smallskip\noindent\textbf{Human Evaluation:} The average annotation scores provided by humans are shown in Table~\ref{tbl:human_prefs}. 
Out of all model families, gpt-3.5-turbo summaries were most preferred with an average score of 3.53 followed by mpt-30b-chat with 3.39 average.
From the different prompt styles we tested, responses generated from TELeR L2 were most preferred with a 3.42 average.
%\santu{briefly describe the results}

\medskip
\noindent\textbf{Automatic Evaluation:}  We report automatic evaluation results for all metrics, all models, and all datasets in Table~\ref{tbl:best_scores}.
This table shows the highest scores achieved by each model across the set of all prompts with different TELeR levels.
For the AllSides dataset, the best-scoring models vary with the evaluation metric used, with some metrics yielding \texttt{phi-3-mini-128k-instruct} as the best, while others favor \texttt{gemini-pro}. For the 3P dataset, \texttt{gpt-3.5-turbo} consistently scored the best with \texttt{gemini-pro} coming in second across most metrics.

% We show the results for automatic evaluation scores in Table \ref{tbl:best_scores}. 
% This table shows the highest scores achieved by each model over the set of all prompts with different TELeR levels.
% We observe that Phi-3-mini-128k-instruct generally performs the best on the AllSides dataset out of all other models, while for the 3P dataset, we see GPT-3.5-turbo on top.

% \santu{What is the point of this paragraph and figure 2? We are not focusing on SEM-F1 anymore, right?}
% In Figure \ref{fig:score_distribution}, we show the distribution of scores assigned by Sem-F1 and by our human annotators for each of our datasets.
% In it, we can see the annotators generally prefer the summaries on the 3P data over the AllSides data.
% This behavior is similarly observed in all the other metrics shown in Table \ref{tbl:best_scores}.

%\input{fig/annotator_agreement}
\begin{figure}[!htb]\vspace{-2mm}
    \centering
    \includegraphics[width=0.9\linewidth]{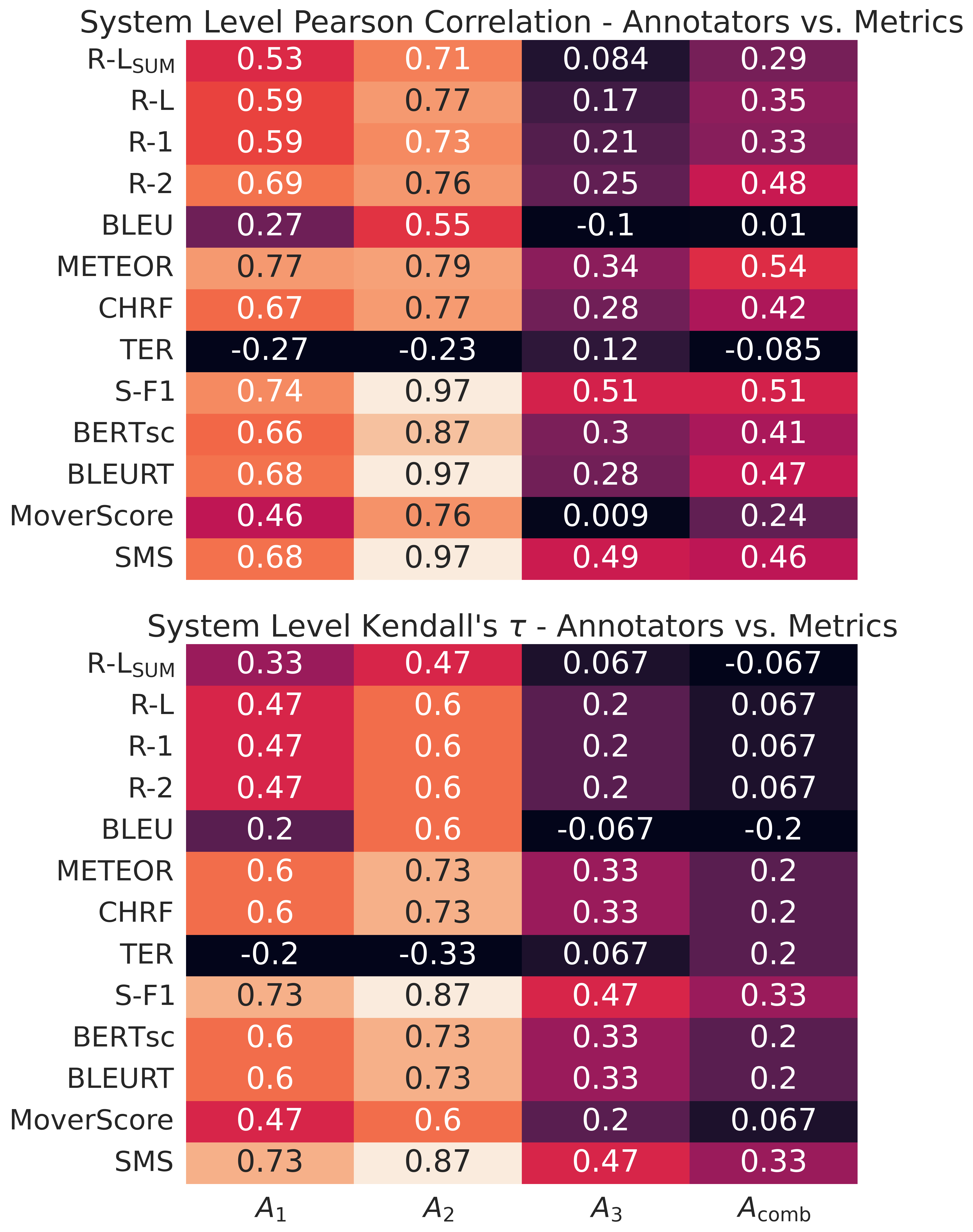}
    \vspace{-3mm}
    \caption{
        System-level Pearson correlation and Kendall's $\tau$ scores between annotator scores and automatic evaluation metrics (higher is better).
        The ``comb'' subscript shows the combined score where the annotators sat with each other to settle on a final score for each annotation sample.
    }
    \label{fig:agreement_system_level}\vspace{-2mm}
\end{figure}

\noindent\textbf{Human Vs.~Automatic Evaluation:} In Figure \ref{fig:agreement_system_level}, we report the System-level Kendall's $\tau$ and Pearson's $\rho$ correlation coefficients between all our metrics and our human annotations \citep{meta_sys_1, meta_sys_2, meta_sys_3, meta_sys_4}.
We show the correlation scores for each individual annotator, but focus on the $A_{\text{comb}}$ field, which represents the final score that was agreed upon by all annotators.
Interestingly, while Sem-F1 was originally proposed as a specialized metric for the SOS task~\citep{semf1} and while this is indeed shown to be the case according to the Kendall's $\tau$ correlation, we can also see that it is matched by SMS and is also seen being beaten by METEOR in Pearson's $\rho$.

\medskip
\noindent\textbf{Key Findings:} Our comprehensive benchmarking study provides us with the following interesting insights regarding the relationships between models, evaluation metrics, TELeR Levels, and human preferences for the SOS task.

\begin{itemize}[leftmargin=*,itemsep=0ex,partopsep=0ex,parsep=0ex]
    \item \textbf{Models vs.~TELeR Levels:} When comparing models against TELeR prompts in Table~\ref{tbl:best_scores}, we found that while TELeR L1 generally perform the best, some models show preferences towards other styles. For example, all the vicuna models show favor over L2 (64 top scores), with much fewer L1 prompts showing top scores (23).
    
    \item \textbf{Datasets vs.~TELeR Levels:} Based on Table \ref{tbl:best_scores}, L1 prompts consistently score the highest, counting 106 and 122 for AllSides and 3P, respectively.
    L2 comes in second place with 49 and 47, suggesting that brevity is preferred in general while designing prompts for the SOS task.
    
    % \item \textbf{Datasets vs.~TELeR Levels:} From Table \ref{tbl:response_lengths}, we again see that L1 prompts yield most of the top scores for both datasets but especially on 3P. 
    % On closer inspection, we also found that behind L1 prompts, the next best results on the AllSides dataset were achieved using L3 prompts.

    % \item \textbf{Automated Metrics Vs.~TELeR Levels:} From Table \ref{tbl:best_scores}, we see that most metrics tend to favor summaries from TELeR L1 prompts.
    
    \item \textbf{Human Preference Vs.~TELeR Levels:} Table \ref{tbl:human_prefs} shows that human annotators showed bias towards TELeR L2 prompts. However, the variance seems to be relatively small across L1 - L4.
\end{itemize}

\section{Conclusion}
In this study, we investigated the capability of LLMs for performing the Semantic Overlap Summarization (SOS) task. 
We evaluated LLMs on an existing dataset and additionally introduced a new dataset called the \textit{PrivacyPolicyPairs} (3P) dataset. To account for the effects of prompt sensitivity, we adopted the TELeR prompting taxonomy to create a diverse set of prompts and found that:
%We use the TELeR prompting taxonomy to devise a set of hand-crafted prompts that generate the highest scores we could achieve with pre-trained instruction-tuned LLMs and found that: 
1) Different TELeR levels impact each model and data set differently, suggesting that the degree of details provided in prompts must be studied and reported before making a final conclusion on LLMs' performance;
2) METEOR, SMS, and Sem-F1 are the metrics that correlate the best with human judgments at the system level; and 
3) Human annotators tend to prefer summaries generated from TELeR L2, i.e., prompts with moderate details.

%3) based on our testing methodology, the best summarization results in a zero-shot setting can be accomplished using \textttt{gpt-3.5-turbo} with TELeR L2 style prompts.
%1) TELeR level $1$  prompts perform the best on average, and 2) generated summaries were more accurate for the Allsides data-set than the 3P data-set in terms of popular evaluation metrics.

% \section{Ethical Considerations}
\section{Limitations}
\noindent\textbf{Dataset Size:}
At only 135 samples, it is not feasible to train a model on just the 3P data alone.
Of course the AllSides dataset exists to accompany the 3P dataset but they represent a different category of documents from the 3P dataset which is another barrier to training.
However while the size of the new dataset is small, there is a large amount of time and resource that is required to build a dataset of this nature.
Firstly, this dataset requires that for each sample, we find two documents that share an overlapping narrative. 
Second, each sample is annotated manually by 3 people which for this dataset results in 405 annotations. 
That is without considering the other annotations where no overlap was found.
Third, there have been several instances where disagreements need to be resolved which requires further discussion among annotators.
Despite these limitations it is worth noting that this work effectively doubles the amount of samples to evaluate on the SOS task when considering both AllSides data and 3P data combined, taking our initial 137 sample news article test set to a combined 272 sample evaluation set over both news articles and privacy policy documents.
In the future, a larger scale effort will be needed to increase the space of data for the SOS task.

\smallskip\noindent\textbf{Human Annotation:}
Annotation work is expensive in both time and money. 
We recruited all our annotators from within our department and saved on money but time cost is unavoidable.
To make the process easier for our volunteers we reduced the amount of annotation samples by selecting 15 samples out of all 272 test set samples between AllSides and 3P.
We also only evaluated the largest/newest models from each model family and finally, we wonly evaluated summaries that correspond to the best-performing prompts within each TELeR level.
It is also important to note that the annotation process was purely for scoring user preference and there is no "right" or "wrong" answers to validate.

Despite the limited number of samples, we believe our human evaluation offers sufficient depth and rigor to support meaningful conclusions.
Specifically:
\begin{enumerate}[leftmargin=*,itemsep=0.2ex,partopsep=-0.8ex,parsep=-0.2ex]% \small
    \item Each summary was independently scored by three annotators, followed by joint adjudication to ensure consistency and resolve disagreements. This consensus-based approach improves annotation quality and mitigates individual biases.
    \item The evaluated samples were carefully selected to balance coverage across domains including 7 paris from AllSides and 8 from 3P. his domain diversity enhances the generalizability of our findings across different types of narrative content.
    \item We focused annotation efforts on the strongest-performing prompts and the most competitive models, concentrating the analysis on realistic and high-quality system outputs. This targeted evaluation ensures that performance comparisons are meaningful and relevant to state-of-the-art LLM usage.
\end{enumerate}

\smallskip\noindent\textbf{Model Finetuning:}
For this work we did not perform any fine-tuning on the evaluated models.
All scores were obtained using the pre-trained weights for each model.
This means that it's possible for additional performance to be gained using methods like LoRA \cite{lora}.
However the main goal of this study was to benchmark LLMs to set new baselines for the SOS task.
In that regard we believe this to be an appropriate setup.

\smallskip\noindent\textbf{Automatic Evaluation:}
In this work we show that automatic evaluation cannot yet be trusted for the SOS task.
However, reporting automatic evaluation metrics is standard practice so it is important that we take precaution when using these values to draw conclusions.

\bibliography{anthology,custom,sofsat}
\bibliographystyle{acl_natbib}

% \clearpage
\appendix

\section{Appendix}
\subsection{Additional Figures}
\label{appendix:figures}
\begin{figure}[!htb]
    \includegraphics[width=\linewidth]{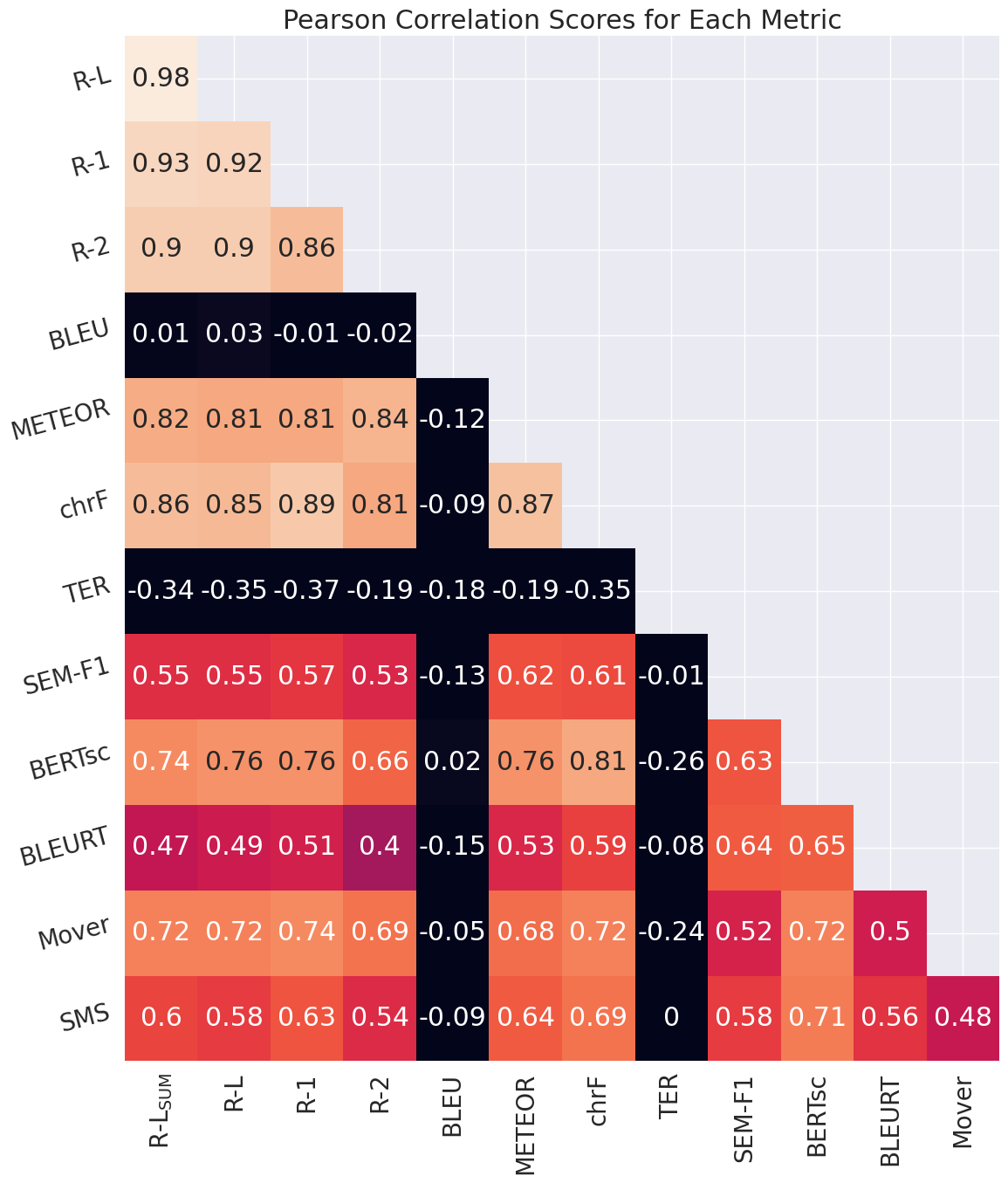}
    \caption{
        Raw correlation scores between all evaluation metrics.
    }
    \label{fig:metric_correlation}
\end{figure}

Figure \ref{fig:metric_correlation} shows Pearson's correlation scores between all metrics on both datasets. 
The Pearson scores were computed using the SciPy library \cite{scipy}

\subsection{More on the 3P Dataset}
In table \ref{tbl:ds_stats}, we show statistics of the 3P dataset.
Figure \ref{tbl:3psample} shows an example of what a sample in the 3P dataset looks like.

% \usepackage{booktabs}
% \usepackage{graphicx}
\begin{table}[!htb]\small
\centering
\begin{tabular}{@{}ll@{}}
\hline
\multicolumn{2}{c}{\textbf{3P Dataset Statistics}} \\
\hline\hline
\# Samples                          & 135       \\
Avg. \# Words per Document          & 331.00    \\
Avg. \# Words per Document Pair     & 662.01    \\
Avg. \# Sentences per Document      & 14.96     \\
Avg. \# Sentences per Document Pair & 28.99     \\
Avg. \# Words per Reference         & 22.46     \\
Avg. \# Sentences per Reference     & 1.75      \\
\hline
\end{tabular}
\vspace{-2mm}
\caption{Dataset statistics for the 3P dataset consisting of 135 document pairs with 3 references each.}\vspace{-2mm}
\label{tbl:ds_stats}
\end{table}
\begin{table*}[!htb]
\centering
\includegraphics[width=.9\linewidth]{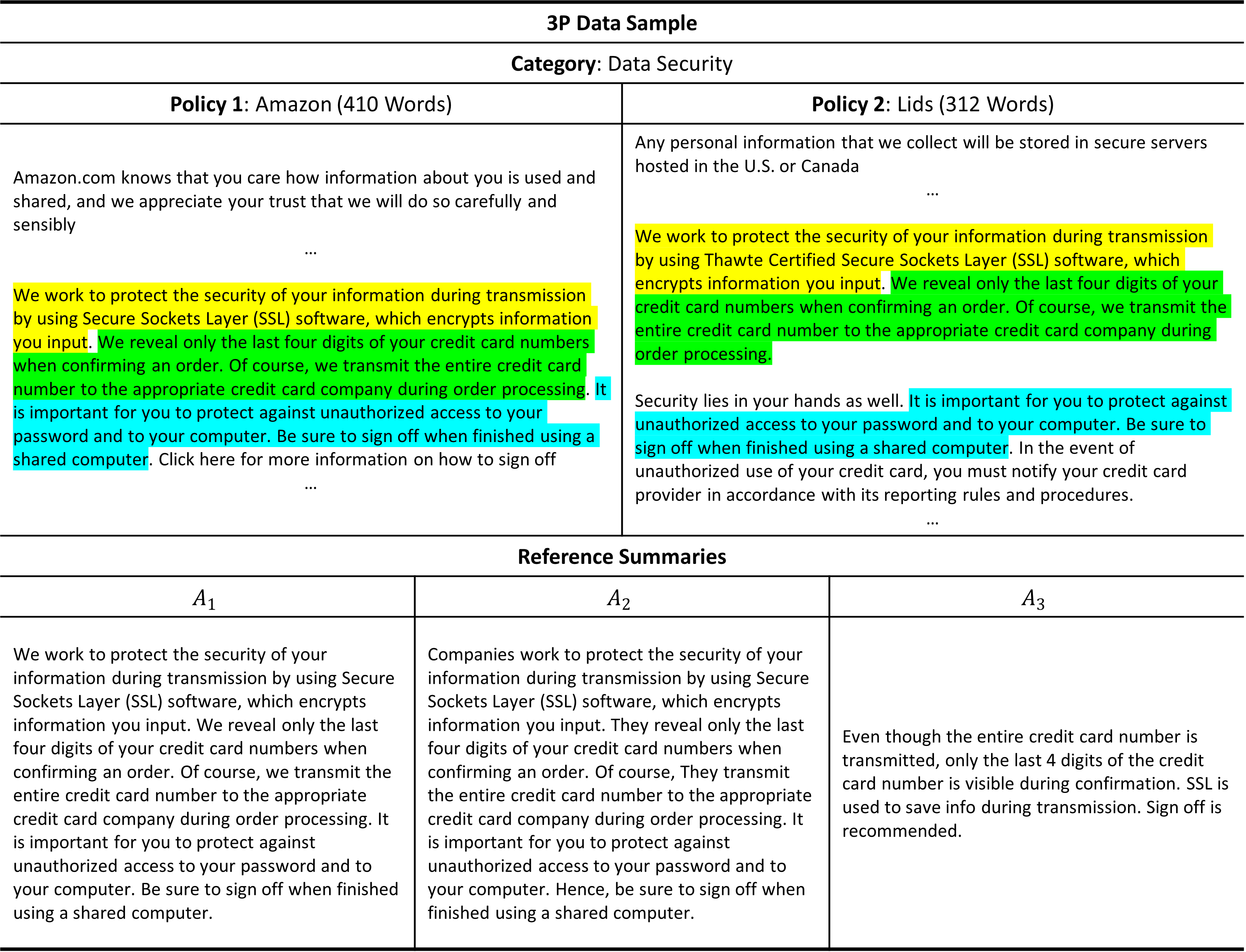}
\vspace{-2mm}
\caption{A single sample from the 3P dataset. For each sample, you are given the category name, company names, the corresponding policy subsections, the count of words in each policy, and the $3$ reference summaries. The highlighted text shows the overlapping information.}
\label{tbl:3psample}
\vspace{-2mm}
\end{table*}

% \subsection{Detailed Results}
% Tables \ref{tbl:alldataallsides} and \ref{tbl:alldataprivacy} contain the full breakdown of evaluation scores for our finalized set of TELeR prompts.

%\input{fig/all_data_allsides}
%\input{fig/all_data_privacypolicy}

%\section{Related Work}
\subsection{Related Work}

% regular
% \cite{salience_alloc}

% multidoc
%  \cite{brio}
\textbf{Text Summarization:}  SOS is essentially a summarization task. Over the past two decades, many document summarization approaches have been investigated~\cite{zhong2019searching}. 
The two most popular among them are \textit{extractive} approaches \citep{cao-etal-2018-retrieve, narayan2018ranking, wu2018learning, zhong2020extractive} and \textit{abstractive} approaches \citep{bae2019summary, liu2017generative, nallapati2016abstractive}. 
Some researchers have tried combining extractive and abstractive approaches \citep{chen2018fast, hsu2018unified, zhang2019pegasus}. 

\smallskip
\noindent\textbf{Semantic Overlap Summarization:}
Semantic Overlap Summarization (SOS) is a task aimed at extracting and condensing shared information between two input documents, $D_A$ and $D_B$. 
The output, denoted as $D_O$, is generated in natural language and only includes information present in both input documents. 
The task is framed as a constrained multi-seq-to-seq (text generation) task, where brevity is emphasized to minimize the repetition of overlapping content. 
The output can be extractive summaries, abstractive summaries, or a combination of both~\cite{sofsat}.
This is similar to the sentence intersection task, where your input is comprised of sentences instead of documents and your output contains only the common information \citep{sent_intersect1,sent_intersect2}.

To facilitate research in this area, \citet{semantic_overlap} introduced the AllSides dataset for training and evaluation, which we also used for evaluation in this work.

\smallskip\noindent\textbf{LLMs and Summarization:}
As the transformer architecture gained popularity, further research showed favorable behavior at scale, allowing the creation of larger and more performant models \citep{scaling1, scaling2, scaling3, scaling4, scaling5}.
With the rising prevalence of these large language models, summarization naturally became one of the many areas of NLP that have progressed as a result. 
LLM performance has been evaluated in tasks such as news summarization \citep{llm_news_summ}, multi-document summarization \citep{llm_multidoc_summ}, and dialogue summarization \citep{dial_summ_1, dial_summ_2} but there has also been research into using them as annotators or evaluators \citep{llm_summ_eval, llm_summ_ann}.

\smallskip\noindent\textbf{Prompt Engineering for LLMs:}  ``Prompt Engineering'' is a technique for maximizing the utility of LLMs in various tasks~\cite{zhou2022large}. It involves crafting and revising the query or context to elicit the desired response or behavior from LLMs~\cite{brown2022does}. Prompt engineering is an iterative process requiring multiple trial and error runs~\cite{shao2023compositional}. In fact, differences in prompts along several key factors can significantly impact the accuracy and performance of LLMs in complex tasks. To address this issue, \citet{teler} recently proposed the TELeR taxonomy, which can serve as a unified standard for benchmarking LLMs' performances by exploring a wide variety of prompts in a structured manner.

%\santu{Show the figure of the taxonomy from the original paper here and describe the taxonomy briefly.}

\label{sec:teler}
\smallskip\noindent\textbf{The TELeR Taxonomy:}
As shown in Figure~\ref{TELeR_Fig}, the TELeR taxonomy introduced by \citet{teler} categorizes complex task prompts based on four criteria. 

\begin{enumerate}[leftmargin=*,itemsep=0.2ex,partopsep=-0.8ex,parsep=-0.2ex]
    
    \item \textbf{Turn}: This refers to the number of turns or shots used while prompting an LLM to accomplish a complex task. In general, prompts can be classified as either single or multi-turn.
    
    \item \textbf{Expression}: This refers to the style of expression for interacting with the LLM, such as questioning or instructing. 
    
    %\item \textbf{Level of Details}: This dimension of prompt style deals with the granularity or depth of question or instruction. Prompts with higher levels of detail provide more granular instructions. For example, Level 0 has no directive, \ie, questions or instructions. On the other hand, Level 4 prompts have to specify the particular sub-tasks, an explanation of the output, and the basis for evaluation of the output generated by the LLM.
    \item \textbf{Level of Details}: This dimension of prompt style deals with the granularity or depth of question or instruction. Prompts with higher levels of detail provide more granular instructions. %For example, level 0 has no directive, \ie, questions or instructions, whereas level 1 prompts do. 
    
    \item \textbf{Role}: LLMs can provide users with the option of specifying the role of the system. The response of LLM can vary due to changes in role definitions in spite of the fact that the prompt content remains unchanged.
\end{enumerate}

The taxonomy outlines $7$ distinct levels starting from level 0 to level 6.
With each increase in level comes an increase in complexity of the prompt.
In level 0, only data/context is provided with no further instruction.
Level 1 extends level 0 by providing single-sentence instruction.
Then level 2 extends level 1, and so on, until level 6, where all characteristics of previous levels are provided along with the additional instruction for the LLM to explain its output. For more details on the TELeR taxonomy and its applications, see~\citet{teler}.
For convenience, we include the outline diagram from the paper in Appendix \ref{appendix:prompt_design}.

%\santu{You need to write a related work section here and connect this work to existing literature.}
\subsection{Evaluation Metrics}
\label{appendix:eval}

\smallskip\noindent\textbf{SEM-F1}~\citep{semf1}: Semantic $\text{F}_1$ computes the sentence-wise similarity (e.g., cosine similarity between two sentence embeddings) to infer the semantic overlap between a system-generated sentence and a reference sentence from both precision and recall perspectives and then, combine them into the F1 score.

\smallskip\noindent\textbf{BERTscore}~\citep{bertscore}: An automatic evaluation metric for text generation. Analogously to common metrics, BERTScore computes a similarity score for each token in the candidate sentence with each token in the reference sentence.

\smallskip\noindent\textbf{ROUGE}~\citep{rouge}: Recall-Oriented Understudy for Gisting Evaluation counts the number of overlapping units such as n-gram, word sequences, and word pairs between the computer-generated summary to be evaluated and the ideal summaries created by humans. This metric is mainly used for evaluating text generation.

\smallskip\noindent\textbf{BLEURT}~\citep{bleurt}: A learned evaluation metric based on BERT that can model human judgments with a few thousand possibly biased training examples. This metric is primarily evaluating machine translation systems.

\smallskip\noindent\textbf{BLEU}~\citep{bleu}: Bilingual Evaluation Understudy score is a precision-based metric that evaluates the quality of generated text by measuring n-gram overlap between the generated and reference texts. It is primarily used for machine-translation tasks.

\smallskip\noindent\textbf{METEOR}~\citep{meteor}: An automatic metric for machine translation evaluation that is based on a generalized concept of unigram matching between the machine-produced translation and human-produced reference translations.

\smallskip\noindent\textbf{chrF}~\citep{chrf}: character n-gram F-score for automatic evaluation of machine translation output.

\smallskip\noindent\textbf{MoverScore}~\citep{moverscore}: Built upon a combination of contextualized representations of system and reference texts and  a distance between these representations measuring the semantic distance between system outputs and references.

\smallskip\noindent\textbf{Sentence Mover's Similarity}~\citep{sms}: Measures the semantic similarity between two texts by computing the minimum cost of transforming one set of sentence embeddings into another using the Earth Mover’s Distance (EMD).

\smallskip\noindent\textbf{CIDEr}~\citep{cider}: Measures the similarity between generated and reference texts by computing TF-IDF-weighted n-gram overlap, emphasizing important and distinctive words. It was originally designed for image captioning

\smallskip\noindent\textbf{TER}~\citep{ter}: Measures the number of edits (insertions, deletions, substitutions, and shifts) needed to transform a generated text into a reference text, normalized by the total number of words in the reference. Lower TER scores indicate better translations, as fewer edits are required.

\subsection{System Level and Summary Level Correlation}
To understand the performance of automatic evaluation metrics in comparison to human evlautions we examine the correlations between the distribution of scores.

Rather than a raw correlation computation between human scores and automatic scores, the system-level and summary-level methods are the commonly used for computing correlation \citep{meta_sys_1, meta_sys_2, meta_sys_3, meta_sys_4}. 

We use the definition from \citet{revisiting_gold} to describe these methods.
Given $m$ system outputs on each of the $n$ data samples and two different evaluation methods (human evaluations vs automatic evaluations) resulting in two $n$-row, $m$-column score matrices $X$ and $Y$ , the summary-level correlation is an average of samplewise correlations:

$$
r_{sum}(X, Y) = \frac{\sum_i\mathcal{C}(X_i, Y_i)}{n},
$$
where $X_i$, $Y_i$ are the evaluation results on the $i$-th data sample and $\mathcal{C}$ is a function calculating a correlation coefficient (\eg, the Pearson correlation coefficient). 
In contrast, the system-level correlation is calculated on the aggregated system scores:

$$
r_{sys}(X, Y) = \mathcal{C}(\bar{X}, \bar{Y}),
$$

where $\bar{X}$ and $\bar{Y}$ contain $m$ entries which are the system scores from the two evaluation methods averaged across $n$ data samples, \eg, $\bar{X}_0 = \sum_i X_{i, 0}/n$

\subsection{Prompt Design}
\label{appendix:prompt_design}

\begin{figure*}[!htb]
  \centering    \includegraphics[width=0.9\textwidth]{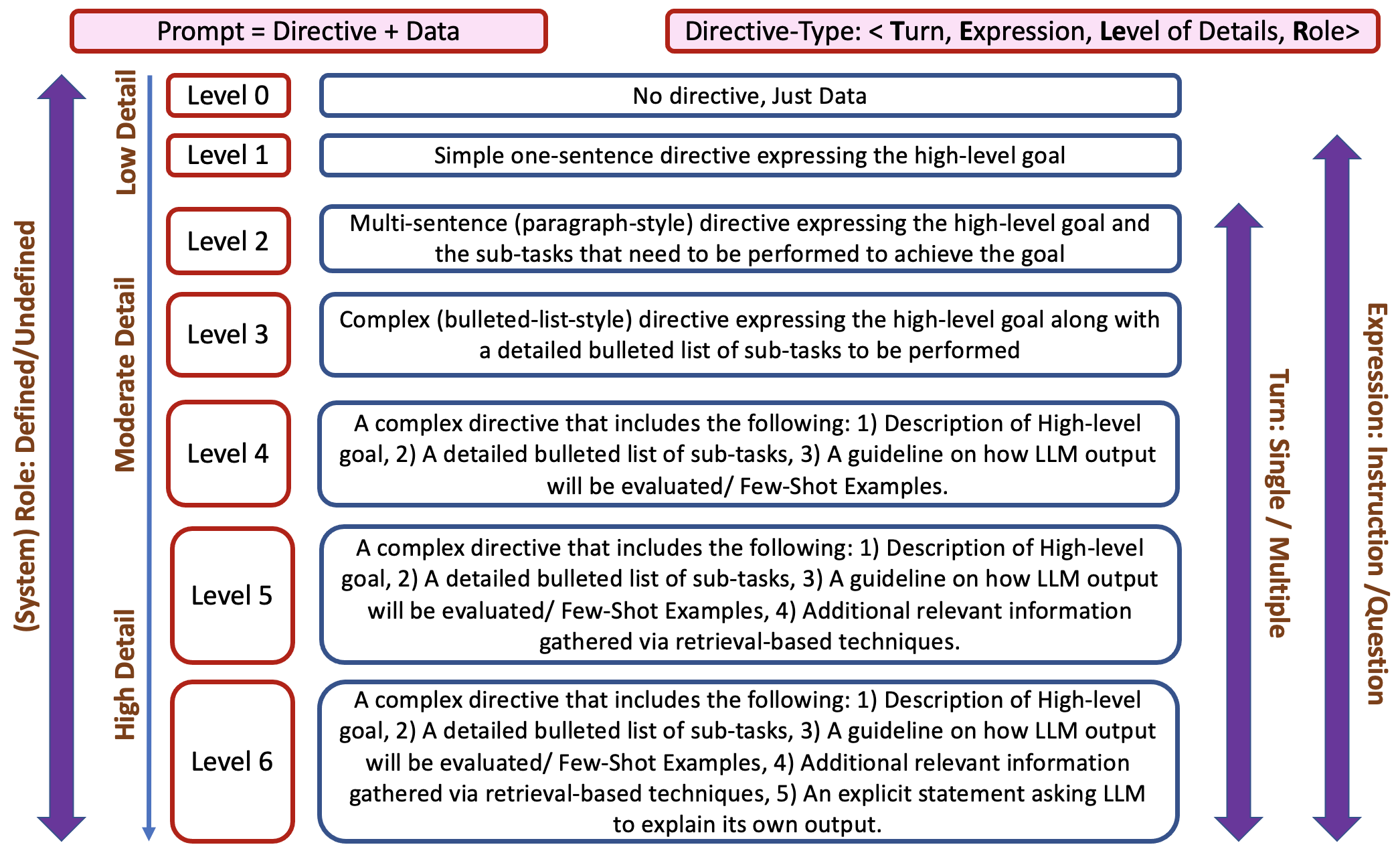}
  \caption{\textbf{TELeR} Taxonomy proposed by \citet{teler}:  (<\textbf{T}urn, \textbf{E}xpression, \textbf{Le}vel of Details, \textbf{R}ole>)}
  \label{TELeR_Fig}
\end{figure*}

We prompted LLMs in a zero-shot setting with TELeR since zero-shot approaches to NLP tasks have gained popularity with the growing capabilities of LLMs. 
For example, works from~\citet{zero_shot_topic_inf, zero_shot_cls} explore their zero-shot use cases in topic inference and text classification.
The taxonomy is best outlined by Figure \ref{TELeR_Fig}.

For this study, we used TELeR levels 0 through 4 (5 out of the 7). 
We chose not to prompt using levels 5 and 6 because their use of retrieval augmented prompting does not necessarily apply to the SOS task.
This is due to all relevant context being present, \ie the two source narratives are already provided as part of the prompt. 
Furthermore, requirement number 5 for level 6 also specifies asking the LLM to explain its own output, which would negatively affect the generated summaries during evaluation. We also experiment with in-context learning prompts~\cite{gpt3}.

In Section \ref{sec:designed_prompts}, we discussed having different prompt variations for TELeR levels 0 through 4 and In-Context Learning prompts.
The number of variations for each group is shown in Table~\ref{tbl:prompt_variations}.

\begin{table}[!htb]
    \resizebox{\linewidth}{!}{%
    \begin{tabular}{lccc|c}
        \hline
        Template Group & For PPP & For AllSides  & For Both & Total   \\
        \hline\hline
        Systm Role          & 2 & 2 & 6 &  10 \\
        TELeR L0            & 0 & 0 & 1 & 1 \\
        TELeR L1            & 3 & 3 & 5 & 11 \\
        TELeR L2            & 3 & 3 & 3 & 9 \\
        TELeR L3            & 3 & 3 & 2 & 8 \\
        TELeR L4            & 3 & 3 & 2 & 8 \\
        In-Context Learning & 0 & 0 & 1 & 1\\
        \hline
    \end{tabular}%
    }\vspace{-2mm}
    \caption{
        The number of prompts created for each template group.
        The "For PPP/AllSides columns indicate how many prompts were created for that dataset only.
        The "For Both" column is for the prompts that could be applied to both datasets.
        For exact prompt details, refer to Appendix \ref{appendix:prompt_design} for exact prompt contents.
    }
    \label{tbl:prompt_variations}

    \vspace{-2mm}
\end{table}

For each group, our templates follow these general patterns:

\begin{itemize}[leftmargin=*,itemsep=0.2ex,partopsep=-0.8ex,parsep=-0.2ex] \small
    \item \textbf{TELeR Level 0}: \{\textbf{Document 1}\} \{\textbf{Document 2}\}
    \item \textbf{TELeR Level 1}: 
        \begin{quote}
            Document 1: \{\textbf{Document 1}\} \\
            Document 2: \{\textbf{Document 2}\}
            
            Summarize the overlapping information between these two documents
        \end{quote}
    \item \textbf{TELeR Level 2}: 
        \begin{quote}
        \{\textbf{TELeR Level 1 Prompt Text}\}
        
        This information must keep in mind the 5W1H facets of the documents. Do not include any uncommon information.
        \end{quote}
    \item \textbf{TELeR Level 3}:
        \begin{quote}
        \{\textbf{TELeR Level 1 Prompt Text}\}
        \begin{itemize}
            \item This information must keep in mind the 5W1H facets of the documents. 
            \item Do not include uncommon information.
        \end{itemize}
        \end{quote}
    \item \textbf{TELeR Level 4}:
        \begin{quote}
        \{\textbf{Level 3 Prompt Text}\}. \\
        Your response will be evaluated against a set of reference summaries. Your score will depend on how semantically similar your response is to the reference.
        \end{quote}
    \item \textbf{In-context Learning}:
        \begin{quote}
        Document 1: \{\textbf{Example Document 1}\} \\
        Document 2: \{\textbf{Example Document 2}\} \\
        Summary: \{\textbf{Example Summary}\} \\
        
        Document 1: \{\textbf{Document1}\} \\
        Document 2: \{\textbf{Document2}\} \\
        Summary:
        \end{quote}
\end{itemize}

The exact prompts are laid out in the following passage.

\smallskip\noindent\textbf{System Role Variations} Our system role templates are made up of 2 AllSides-specific items, 2 3P specific-items and 6 for general purpose. These are written as follows
\begin{itemize}[leftmargin=*,itemsep=0.2ex,partopsep=0ex,parsep=0ex] \small
    \item\textbf{AllSides}
    \begin{itemize}
        \item you will be given two news articles to read. then you will be given an instruction. follow these instructions as closely as possible
        \item you will read two news articles and answer any questions about them
    \end{itemize}
    \item\textbf{3P}
    \begin{itemize}
        \item you are to read two privacy policies and briefly provide information according to the user's needs
        \item you are to read two privacy policies and provide concise answers to the user
    \end{itemize}
    \item\textbf{Both}
    \begin{itemize}
      \item you are to read several documents and briefly provide information according to the user's needs
      \item you are to read several documents and provide concise answers to the user
      \item you will read two documents and give brief answers to user questions
      \item you are a machine who is given 3 inputs: document 1, document 2, and the instructions. your output will adhere to these 3 inputs.
      \item you will be given 2 documents and a set of instructions. follow the instructions as closely as possible.
      \item you will be given 2 documents and a set of instructions. your response to these instructions will rely on the material covered in the 2 documents.
    \end{itemize}
\end{itemize}

\smallskip\noindent\textbf{In-Context Learning Template}: We use the following for our in-context learning template:
\begin{itemize}[leftmargin=*,itemsep=0.2ex,partopsep=0ex,parsep=0ex] \small
    \item 
    \begin{quote}
    Document 1:
    \textbf{\{\{Example Document 1\}\}}

    Document 2:
    \textbf{\{\{Example Document 2\}\}}
    
    Summary:
    \textbf{\{\{Example Reference\}\}}\\

    Document 1:
    \textbf{\{\{Document 1\}\}}

    Document 2:
    \textbf{\{\{Document 2\}\}}

    Summary:
    \end{quote}
\end{itemize}

\smallskip\noindent\textbf{TELeR Level 0 Template}: With no possibility for variation, our TELeR L0 template is written as follows:
\begin{itemize}[leftmargin=*,itemsep=0.2ex,partopsep=0ex,parsep=0ex] \small
    \item \{\textbf{Document 1}\} \{\textbf{Document 2}\}
\end{itemize}

\smallskip\noindent\textbf{TELeR Level 1 Template}: For our TELeR L1 templates we have 3 AllSides-only items, 3 3P-only items, and 5 general-purpose items.
\begin{itemize}[leftmargin=*,itemsep=0.2ex,partopsep=0ex,parsep=0ex] \small
    \item\textbf{AllSides}
        \begin{itemize}
        \item
        Document 1:
        \textbf{\{\{Document 1\}\}}

        Document 2:
        \textbf{\{\{Document 2\}\}} \\

        In one sentence, please tell me the overlapping information between article 1 and article 2
    
        \item
        Document 1:
        \textbf{\{\{Document 1\}\}}

        Document 2:
        \textbf{\{\{Document 2\}\}} \\

        summarize the overlapping information between the articles
    
        \item
        Document 1:
        \textbf{\{\{Document 1\}\}}

        Document 2:
        \textbf{\{\{Document 2\}\}} \\

        output the overlapping information of the events covered in these articles
        \end{itemize}

    \item\textbf{3P}
        \begin{itemize}
        \item
        Policy 1:
        \textbf{\{\{Document 1\}\}}

        Policy 2:
        \textbf{\{\{Document 2\}\}} \\

        In one sentence, please tell me the overlapping information between policy 1 and policy 2
    
        \item
        Policy 1:
        \textbf{\{\{Document 1\}\}}

        Policy 2:
        \textbf{\{\{Document 2\}\}} \\

        summarize the information that the two policies share
    
        \item
        Policy 1:
        \textbf{\{\{Document 1\}\}}

        Policy 2:
        \textbf{\{\{Document 2\}\}} \\

        what is the shared information between the two policies
        \end{itemize}

    \item\textbf{Both}
    \begin{itemize}
        \item
        Document 1:
        \textbf{\{\{Document 1\}\}}

        Document 2:
        \textbf{\{\{Document 2\}\}} \\

        In one sentence, please tell me the overlapping information between Document 1 and Document 2
    
        \item
        Document 1:
        \textbf{\{\{Document 1\}\}}

        Document 2:
        \textbf{\{\{Document 2\}\}} \\

        summarize the overlapping information between the documents.
    
        \item
        Document 1:
        \textbf{\{\{Document 1\}\}}

        Document 2:
        \textbf{\{\{Document 2\}\}} \\

        output the overlapping information between the documents.
    
        \item
        Document 1:
        \textbf{\{\{Document 1\}\}}

        Document 2:
        \textbf{\{\{Document 2\}\}} \\

        output the common information between the documents.
    
        \item
        Document 1:
        \textbf{\{\{Document 1\}\}}

        Document 2:
        \textbf{\{\{Document 2\}\}} \\

        output only the overlapping information
    \end{itemize}
    
\end{itemize}

\smallskip\noindent\textbf{TELeR Level 2 Variations}: For our TELeR L2 templates we have 3 AllSides-only items, 3 3P-only items, and 3 general-purpose items.
\begin{itemize}[leftmargin=*,itemsep=0.2ex,partopsep=0ex,parsep=0ex] \small
    \item\textbf{AllSides}
        \begin{itemize}
            \item
            Document 1:
            \textbf{\{\{Document 1\}\}}

            Document 2:
            \textbf{\{\{Document 2\}\}} \\
        
            these articles share similarities. output the information that is shared between them. keep your output short. to be as accurate as possible, cover the "who, what, when, where, and why of the shared information.
        
            \item
            Document 1:
            \textbf{\{\{Document 1\}\}}

            Document 2:
            \textbf{\{\{Document 2\}\}} \\
        
            who or what are the common subjects of the two documents? what events are common between the documents? do the documents mention any locations that are the same between the two? give your response in a single sentence.
        
            \item
            Document 1:
            \textbf{\{\{Document 1\}\}}

            Document 2:
            \textbf{\{\{Document 2\}\}} \\
        
            summarize the overlap
        \end{itemize}
    
    \item\textbf{3P}
        \begin{itemize}
            \item
            Policy 1:
            \textbf{\{\{Document 1\}\}}

            Policy 2:
            \textbf{\{\{Document 2\}\}} \\

            These policies are categorized under "{{Category}}". Describe the common aspects of these two policies in terms of this category. make sure to include the shared entities, actions and scope of the documents. Do not make any mention of information that is not shared between them. Keep your response short
        
            \item
            Policy 1:
            \textbf{\{\{Document 1\}\}}

            Policy 2:
            \textbf{\{\{Document 2\}\}} \\

            These policies are categorized under "{{Category}}". Describe the common aspects of these two policies in terms of this category. make sure to include the shared entities, actions and scope of the documents. Do not make any mention of information that is not shared between them. give your response in a single sentence.
        
            \item
            Policy 1:
            \textbf{\{\{Document 1\}\}}

            Policy 2:
            \textbf{\{\{Document 2\}\}} \\

            These privacy policy excerpts are tagged with the category: "{{Category}}". summarize the overlapping information between the documents. to be as accurate as possible, cover the who, what, when, where, and why of the common information.
        \end{itemize}

    \item\textbf{Both}
        \begin{itemize}
            \item
            Document 1:
            \textbf{\{\{Document 1\}\}}

            Document 2:
            \textbf{\{\{Document 2\}\}} \\
        
            summarize the overlapping information between the two documents. explain the who, what, when, where, and why to give full context. 
        
            \item
            Document 1:
            \textbf{\{\{Document 1\}\}}

            Document 2:
            \textbf{\{\{Document 2\}\}} \\
        
            summarize the overlapping information between the two documents. explain the who, what, when, where, and why to give full context. the output should be two sentences at most.
        
            \item
            Document 1:
            \textbf{\{\{Document 1\}\}}

            Document 2:
            \textbf{\{\{Document 2\}\}} \\
        
            output the shared information between the documents. do not include any information outside of the shared information. keep your response short.
        \end{itemize}
    
\end{itemize}

\smallskip\noindent\textbf{TELeR Level 3 Variations}: For our TELeR L3 templates we have 3 AllSides-only items, 3 3P-only items, and 2 general-purpose items.
\begin{itemize}[leftmargin=*,itemsep=0.2ex,partopsep=0ex,parsep=0ex] \small
    \item\textbf{AllSides}
        \begin{itemize}
            \item
            Document 1:
            \textbf{\{\{Document 1\}\}}

            Document 2:
            \textbf{\{\{Document 2\}\}} \\

            please answer the following:\\
            - who or what are the common subjects of the two documents\\
            - what events are common between the documents\\
            - do the documents mention any locations that are the same between the two\\
            - keep your response brief. 2 sentences max.
            \item
            Document 1:
            \textbf{\{\{Document 1\}\}}

            Document 2:
            \textbf{\{\{Document 2\}\}} \\

            Consider the following questions and respond in a single sentence:\\
            - who or what are the common subjects of the two documents\\
            - what events are common between the documents\\
            - do the documents mention any locations that are the same between the two
        \end{itemize}
    
    \item\textbf{3P}
        \begin{itemize}
            \item
            Policy 1:
            \textbf{\{\{Document 1\}\}}

            Policy 2:
            \textbf{\{\{Document 2\}\}} \\

            These policies are categorized under "{{Category}}". With this in mind, please answer the following:\\
            - Describe the common aspects of these two policies in terms of this category.\\
            - make sure to include the shared entities, actions and scope of the documents.\\
            - Do not make any mention of information that is not shared between them.\\
            - Do not respond in a list format and instead respond normally.\\
            - Keep your response to 3 sentences at most
        
            \item
            Policy 1:
            \textbf{\{\{Document 1\}\}}

            Policy 2:
            \textbf{\{\{Document 2\}\}} \\

            These policies are labelled under the "{{Category}}" category. With this in mind, use a single sentence that answers the following:\\
            - Describe the common aspects of these two policies in terms of this category.\\
            - make sure to include the shared entities, actions and scope of the documents.\\
            - Do not make any mention of information that is not shared between them.\\
            - Do not respond in a list format and instead respond normally.
        
            \item
            Policy 1:
            \textbf{\{\{Document 1\}\}}

            Policy 2:
            \textbf{\{\{Document 2\}\}} \\

            These policies are labelled under the "{{Category}}" category. With this in mind, use a single sentence that answers the following:\\
            - summarize the information that is shared between the policies\\
            - cover the who, what, when, where, and why of the common information\\
            - respond in as few sentences as possible
        \end{itemize}

    \item\textbf{Both}
        \begin{itemize}
            \item
            Document 1:
            \textbf{\{\{Document 1\}\}}

            Document 2:
            \textbf{\{\{Document 2\}\}} \\

            please answer the following:\\
            - who or what are the common subjects of the two documents\\
            - what events are common between the documents\\
            - do the documents mention any locations that are the same between the two\\
            - keep your response brief. 2 sentences max.
            \item
            Document 1:
            \textbf{\{\{Document 1\}\}}

            Document 2:
            \textbf{\{\{Document 2\}\}} \\

            Consider the following questions and respond in a single sentence:\\
            - who or what are the common subjects of the two documents\\
            - what events are common between the documents\\
            - do the documents mention any locations that are the same between the two
        \end{itemize}
    
\end{itemize}

%TODO
\smallskip\noindent\textbf{TELeR Level 4 Variations} For our TELeR L4 templates we have 3 AllSides-only items, 3 3P-only items, and 2 general-purpose items.
\begin{itemize}[leftmargin=*,itemsep=0.2ex,partopsep=0ex,parsep=0ex] \small
    \item\textbf{AllSides}
        \begin{itemize}
            \item
            Document 1:
            \textbf{\{\{Document 1\}\}}

            Document 2:
            \textbf{\{\{Document 2\}\}} \\

            your goal is to describe all the common information between the given documents. to accomplish this you will need to answer the following:\\
            - who or what are the common subjects of the two documents\\
            - what events are common between the documents\\
            - do the documents mention any locations that are the same between the two\\
            - keep your response brief. 2 sentences max.\\
        
            For Example:\\
            Doc1: i have a dog. it's pretty fast.\\
            Doc2: i have a dog. he is a slow runner\\
            Reference Summary: i have a dog.
            \item
            Document 1:
            \textbf{\{\{Document 1\}\}}

            Document 2:
            \textbf{\{\{Document 2\}\}} \\

            your goal is to describe all the common information between the given documents. to accomplish this you will need to answer the following:\\
            - who or what are the common subjects of the two documents\\
            - what events are common between the documents\\
            - do the documents mention any locations that are the same between the two\\
        
            your response will be evaluated according to how similar it is to a "reference summary".\\
            Example:\\
            Question: what is common between the sentence "the dog is slow" and "the dog is fast"\\
            Reference Summary: Both sentences talk about the speed of a dog
            \item
            Document 1:
            \textbf{\{\{Document 1\}\}}

            Document 2:
            \textbf{\{\{Document 2\}\}} \\

            your goal is to describe all the common information between the given documents in one sentence. your single-sentence response will need to capture the following:\\
            - the common events\\
            - common people\\
            - common locations\\
            - the overlapping narrative of the documents\\
        
            your response will be evaluated according to how similar it is to a "reference summary".\\
            Example:\\
            Doc1: the dog is slow\\
            Doc2: the dog is fast\\
            Reference Summary: Both sentences talk about the speed of a dog
        \end{itemize}
    
    \item\textbf{3P}
        \begin{itemize}
            \item
            Policy 1:
            \textbf{\{\{Document 1\}\}}

            Policy 2:
            \textbf{\{\{Document 2\}\}} \\

            your goal is to describe all the common information between the given privacy policies. 
            to accomplish this you will need to answer according to the following:\\
            - Describe the common aspects of these two policies in terms of this category.\\
            - make sure to include the shared entities, actions and scope of the documents.\\
            - Do not make any mention of information that is not shared between them.\\
            - Do not respond in a list format and instead respond normally.\\
            - Keep your response to 3 sentences at most\\
        
            your response will be evaluated according to how similar it is to a "reference summary".\\
            For example, an output of "cat" could be compared to "light" to get a score of 0 but that same output could be compared to "cat" to receive a score of 100. These reference summaries are usually quite short so it is important to keep your response to 3 sentences or less.\\
        
            your response will be evaluated according to how similar it is to a "reference summary".
            Example:\\
            Doc1: the dog is slow\\
            Doc2: the dog is fast\\
            Reference Summary: Both sentences talk about the speed of a dog
            \item
            Policy 1:
            \textbf{\{\{Document 1\}\}}

            Policy 2:
            \textbf{\{\{Document 2\}\}} \\

            your goal is to describe all the common information between the given documents in one sentence. your single-sentence response will need to include the following:\\
            - common aspects related to the given category\\
            - common entities\\
            - common applications\\
        
            your response will be evaluated according to how similar it is to a "reference summary".\\
        
            Example Documents:\\
            Doc1: the dog is slow\\
            Doc2: the dog is fast\\
        
            Example Response:\\
            Both sentences talk about the speed of a dog
        
            \item
            Policy 1:
            \textbf{\{\{Document 1\}\}}

            Policy 2:
            \textbf{\{\{Document 2\}\}} \\

            your goal is to describe all the common information between the given documents in one sentence. your single-sentence response will need to include the following:\\
            - common aspects related to the given category\\
            - common entities\\
            - common applications\\
        
            your response will be evaluated according to how similar it is to a "reference summary".\\
        
            Example Documents:\\
            Doc1: the dog is slow\\
            Doc2: the dog is fast\\
        
            Example Response:\\
            Both sentences talk about the speed of a dog
        \end{itemize}

    \item\textbf{Both}
        \begin{itemize}
            \item
            Document 1:
            \textbf{\{\{Document 1\}\}}

            Document 2:
            \textbf{\{\{Document 2\}\}} \\

            Write a summary of the given documents that follows these instructions:\\
            - who or what are the common subjects of the two documents\\
            - what events are common between the documents\\
            - do the documents mention any locations that are the same between the two\\
            - keep your response brief. 2 sentences max.\\
        
            your response will be evaluated according to how similar it is to a "reference summary".\\
            For Example:\\
            Doc1: i have a dog. it's pretty fast.\\
            Doc2: i have a dog. he is a slow runner\\
            Reference Summary: i have a dog.
            \item
            Document 1:
            \textbf{\{\{Document 1\}\}}

            Document 2:
            \textbf{\{\{Document 2\}\}} \\

            Summarize the overlapping information between these documents. your summary should follow these instructions:\\
            - exclude any information that is similar but differing or contradictory\\
            - write the summary as if you were summarizing a single document.\\
            - your summary should be short. keep it within 2 sentences.\\
        
            your response will be evaluated according to how similar it is to a "reference summary".\\
            For Example:\\
            Doc1: i have a dog. it's pretty fast.\\
            Doc2: i have a dog. he is a slow runner\\
            Reference Summary: i have a dog.
        \end{itemize}
    
\end{itemize}

\label{human_annotation}
\subsection{Annotation Details}
\smallskip\noindent\textbf{3P Dataset Annotations}
When constructing the 3P dataset, annotators were instructed as follows:
\begin{quote}
1) You are given a list of document pairs. For each document pair, read and understand the overlapping information between doc1 and doc2.

2) Write a summary that only includes the overlapping information you have identified.

What is overlapping information?
Any information, statement, or fact that is shared between two or more documents
example: 'John doe is on a trip to Las Vegas' and 'John Doe went to see the fight in Vegas' shares the information 'John Doe is in Las Vegas'

What DOES NOT qualify as overlapping information:
shared mentioning of names
example: 'John Doe is a pilot ' and 'John Doe has never been to Canada' does not have any overlapping information
\end{quote}

\smallskip\noindent\textbf{Model Summary Annotations}
As covered in Section \ref{sec:human_evaluation}, we chose our human evaluation samples by 1) evaluating a subset of data that correspond to 15 samples (7 from AllSides and 8 from 3P) out of the 272 test set samples between AllSides and 3P), 2) evaluating only the largest/newest models from each model family, and 3) evaluating only the summaries that correspond to the best performing prompts within each TELeR level. 
To clarify point 3, each TELeR level has a set of templates, as shown in Table \ref{tbl:prompt_variations}. 
TELeR L1, for example, has 8 prompt and 8 system role templates that can be used to prompt the models on the AllSides dataset. 
All possible combinations for TELeR L1 prompt and system role templates give us 64 unique prompts to be applied to the entire dataset.
After collecting responses and evaluating the average performance for each of the 64 unique prompts, the samples associated with the prompt that yielded the best performance over the AllSides dataset were chosen for human annotation.

When evaluating the summaries generated by the LLMs, annotators were instructed as follows:
\begin{quote}
1) You are given a list of document pairs. For each document pair, read and understand the overlapping information between doc1 and doc2.

3) Read each of the corresponding 'response' entries and assign a score between 0 and 5 (decimal values indluded) based on how well you think it covers the overlapping information
* decimal values such as 1.23 are acceptable scores.

What is overlapping information?
Any information, statement, or fact that is shared between two or more documents
example: 'John doe is on a trip to Las Vegas' and 'John Doe went to see the fight in Vegas' shares the information 'John Doe is in Las Vegas'

What DOES NOT qualify as overlapping information:
shared mentioning of names
example: 'John Doe is a pilot ' and 'John Doe has never been to Canada' does not have any overlapping information
\end{quote}

\subsection{Additional Results}

% \usepackage{multirow}
% \usepackage{graphicx}
\begin{table*}[!htb]\small
\centering
\resizebox{\linewidth}{!}{%
\begin{tabular}{c|c|c|c|c|c|c|c|c|c|c|c|c|c|c}
\hline
%\multicolumn{1}{l|}{\textbf{Dataset}} &
\textbf{Dataset} &
  \textbf{Tmplt.} &
  \textbf{R-L Sum} &
  \textbf{R-L} &
  \textbf{R-1} &
  \textbf{R-2} &
  \textbf{BLEU} &
  \textbf{METEOR} &
  \textbf{chrF} &
  \textbf{TER} $\downarrow$ &
  \textbf{S-F1} &
  \textbf{BERTsc} &
  \textbf{BLEURT} &
  \textbf{MoverScore} &
  \textbf{SMS} \\ \hline\hline
\multirow{6}{*}{AllSides} & L0  & 0.212 & 0.192 & 0.279 & 0.135 & 0.0009 & 0.337 & 36.115 & 1353.976 & 0.476 & 0.173 & -0.637 & 0.548 & 0.546\\
                          & L1  & \textbf{0.276} & 0.258 & 0.356 & \textbf{0.188} & 0.0010 & \textbf{0.407} & 42.538 & 833.364 & \textbf{0.524} & 0.281 & \textbf{-0.474} & 0.568 & 0.561\\
                          & L2  & 0.257 & 0.243 & 0.339 & 0.170 & 0.0010 & 0.386 & 40.701 & 827.023 & 0.516 & 0.240 & -0.558 & 0.562 & 0.549\\
                          & L3  & 0.273 & \textbf{0.263} & \textbf{0.358} & 0.175 & 0.0012 & 0.406 & \textbf{42.696} & 590.499 & 0.499 & \textbf{0.297} & -0.505 & \textbf{0.569} & \textbf{0.565}\\
                          & L4  & 0.259 & 0.250 & 0.335 & 0.162 & \textbf{0.0015} & 0.372 & 39.775 & \textbf{514.080} & 0.457 & 0.244 & -0.646 & 0.561 & 0.548\\
                          & ICL & 0.214 & 0.202 & 0.286 & 0.129 & 0.0010 & 0.342 & 36.837 & 942.628 & 0.423 & 0.179 & -0.768 & 0.543 & 0.542\\ \hline\hline
\multirow{6}{*}{\begin{tabular}[c]{@{}c@{}}Privacy\\ Policy\\ Pairs (3P)\end{tabular}} 
                        & L0  & 0.109 & 0.096 & 0.134 & 0.042 & 0.0008 & 0.218 & 22.929 & 2243.971 & 0.412 & -0.004 & -0.682 & 0.520 & 0.510\\
                        & L1  & \textbf{0.157} & 0.147 & \textbf{0.199} & \textbf{0.062} & 0.0011 & \textbf{0.265} & 30.684 & 1057.247 & \textbf{0.440} & \textbf{0.116} & \textbf{-0.545} & 0.534 & \textbf{0.518}\\
                        & L2  & 0.145 & 0.136 & 0.188 & 0.053 & 0.0008 & 0.254 & 29.823 & 1130.120 & 0.441 & 0.085 & -0.605 & 0.531 & 0.515\\ 
                        & L3  & 0.151 & 0.145 & \textbf{0.199} & 0.048 & 0.0011 & 0.248 & 31.943 & 700.396 & 0.413 & 0.112 & -0.599 & 0.532 & 0.513\\
                        & L4  & 0.152 & \textbf{0.148} & \textbf{0.199} & 0.049 & \textbf{0.0015} & 0.237 & \textbf{30.729} & \textbf{590.374} & 0.393 & 0.104 & -0.661 & 0.529 & 0.505\\
                        & ICL & 0.120 & 0.112 & 0.155 & 0.042 & 0.0010 & 0.219 & 25.154 & 1198.308 & 0.389 & 0.059 & -0.715 & \textbf{0.561} & 0.477\\ \hline
\end{tabular}%
}\vspace{-2mm}
\caption{
    Average scores per metric broken down by level and dataset. 
    Higher is better for all metrics except TER which is denoted by the $\downarrow$.
    TELeR Levels are denoted by "Lx" and In-Context Learning is denoted by "ICL".
    The best of each metric and dataset are in bold. 
    %The bottom row shows the difference between the best scores on the AllSides data and the 3P data along each column.
}
\label{tbl:avg_scores_per_lv}\vspace{-2mm}
\end{table*}

\smallskip\noindent\textbf{Human Preference on Model and Template:}  
While Table \ref{tbl:avg_scores_per_lv} shows that the automatic evaluations tend to have a preference towards TELeR L1 prompts, Table \ref{tbl:human_prefs} shows that human annotators actually tend to prefer TELeR L2 prompts instead.
However, this preference is only 0.04 points ahead of the next best.
The table also indicates the annotators' preference towards \textttt{gpt-3.5-turbo} for the commercial LLMs.
Then, for the open-source LLMs, \textttt{mpt-30b-chat} was the most preferred, with an average annotator score of 3.39.
However, it is important to note that \textttt{Phi-3-mini-128k-instruct} and \textttt{Mistral-7B-Instruct-v0.2} match and beat \textttt{gemini-pro}, respectively, according to humans.

\end{document}